\documentclass{article}
\usepackage{spconf,amsmath,graphicx}

\usepackage{amsmath,amsthm,amsfonts,amssymb}
\usepackage{enumerate}
\usepackage{xcolor}
\usepackage{graphicx}
\usepackage{listings}
\usepackage{float}
\usepackage{algorithm}
\usepackage{algorithmic}
\usepackage{caption}
\usepackage{subfig}
\usepackage{booktabs}

\newcommand{\beq}{\vspace{0mm}\begin{equation}}
\newcommand{\eeq}{\vspace{0mm}\end{equation}}
\newcommand{\beqs}{\vspace{0mm}\begin{eqnarray}}
\newcommand{\eeqs}{\vspace{0mm}\end{eqnarray}}
\newcommand{\barr}{\begin{array}}
\newcommand{\earr}{\end{array}}

\newcommand{\Dmat}{{\bf D}}

\newcommand{\Umat}[0]{{{\bf U}}}

\newcommand{\Wmat}[0]{{{\bf W}}}

\newcommand{\bv}[0]{{\boldsymbol{b}}}
\newcommand{\cv}[0]{{\boldsymbol{c}}}

\newcommand{\fv}[0]{{\boldsymbol{f}}}

\newcommand{\hv}[0]{{\boldsymbol{h}}}
\newcommand{\iv}[0]{{\boldsymbol{i}}}

\newcommand{\ov}[0]{{\boldsymbol{o}}}

\newcommand{\qv}[0]{{\boldsymbol{q}}}

\newcommand{\xv}{\boldsymbol{x}}
\newcommand{\yv}{\boldsymbol{y}}

\newcommand{\thetav}{\boldsymbol{\theta}}

\title{Character-level Deep Conflation for Business Data Analytics}
\name{Zhe Gan$^\dagger$, P. D. Singh$^*$, Ameet Joshi$^\star$, Xiaodong He$^*$, Jianshu Chen$^*$, Jianfeng Gao$^*$, and Li Deng$^*$\thanks{Emails: zhe.gan@duke.edu, \{prabhs, ameetj, xiaohe, jianshuc, jfgao, deng\}@microsoft.com}}
\address{
$^\dagger$Department of Electrical and Computer Engineering, Duke University, Durham, NC\\
$^*$Microsoft Research, Redmond, WA\\
$^\star$Microsoft Corporation, Redmond, WA
}
\begin{document}
\ninept
\maketitle
\begin{abstract}

Connecting different text attributes associated with the same entity (conflation) is important in business data analytics since it could help merge two different tables in a database to provide a more comprehensive profile of an entity. However, the conflation task is challenging because two text strings that describe the same entity could be quite different from each other for reasons such as misspelling. It is therefore critical to develop a conflation model that is able to truly understand the semantic meaning of the strings and match them at the semantic level. To this end, we develop a character-level deep conflation model that encodes the input text strings from character level into finite dimension feature vectors, which are then used to compute the cosine similarity between the text strings. The model is trained in an end-to-end manner using back propagation and stochastic gradient descent to maximize the likelihood of the correct association. Specifically, we propose two variants of the deep conflation model, based on long-short-term memory (LSTM) recurrent neural network (RNN) and convolutional neural network (CNN), respectively. Both models perform well on a real-world business analytics dataset and significantly outperform the baseline bag-of-character (BoC) model.

\end{abstract}
\section{Introduction}
\label{sec:intro}

In business data analytics, different fields and attributes related to the same entities (e.g., same person) are stored in different tables in a database or across different databases. It is important to connect these attributes so that we can get a more comprehensive and richer profile of the entity. This is important because exploiting a more comprehensive profile could lead to better prediction in business data analytics. 

Specifically, the conflation of data aims to connect two rows from the same or different datasets that contain one or more common fields, when the values of the common fields match within a predefined threshold. For example, in the business data considered in this paper, we aim to detect whether two names refer to the same person or not --- see the example in Table \ref{Table:dataset}. Row A and row B represent two name fields from different tables in a database, which is a text string consisting of characters. The strings in the same column of Table \ref{Table:dataset} represent the names of a same person. There are several reasons for the strings in A and B being different: (\emph{i}) possible mis-spelling typos; (\emph{ii}) the lack of suffix; (\emph{iii}) the reverse of family names and given names. Due to these variations and imperfection in data entries, plain keyword matching does not work well~\cite{torvik2009author,milojevic2013accuracy}, and we need a data conflation model in the \emph{semantic} level; that is, the model should be able to identify two different character strings to be associated with a same entity.

To address the aforementioned challenges, we propose character-level deep conflation models that take the raw text strings as the input and predict whether two data entries refer to the same entity. The proposed model consists of two parts: (\emph{i}) a deep feature extractor, and (\emph{ii}) a ranker. The feature extractor takes the raw text string at the character level and produce a finite dimension representation of the text. In particular, we constructed two different deep architectures of feature extractors: (\emph{i}) long-short-term-memory (LSTM) recurrent neural network (RNN) \cite{hochreiter1997long,mikolov2010recurrent} and (\emph{ii}) deep convolutional neural network (CNN) \cite{kim2014convolutional,kalchbrenner2014convolutional}. 
Both deep architectures are able to retain the order information in the input text and extract high-level features from raw data, as shown their great success in different machine learning tasks, including text classification \cite{kim2014convolutional,dai2015semi}, machine translation \cite{kalchbrenner2013recurrent,cho2014learning,sutskever2014sequence,meng2015encoding} and information retrieval \cite{huang2013learning,shen2014latent,palangi2016deep}. Furthermore, extracting the features from the \emph{character}-level is critical in many of the recent success in applying deep learning to natural language processing \cite{kim2015character,ling2015finding,zhang2015character,golub2016character,chung2016character}. As we will show later, our proposed deep conflation model achieves high prediction accuracy in the conflation task for business data, and greatly outperform strong baselines.

\begin{table}[t]
	\caption{\small{Example text string pairs in the dataset.} }\label{Table:dataset}
	\centering
	\small
	\begin{tabular}{l|l|l|l}
		\toprule
		\textbf{A} & emilio yentsch & enrique hafner & javier creswell \\
		\midrule
		\textbf{B} & ydntsch emilip & Mr. halner exrique & Prof. crrxwell javzfr \\
		\bottomrule
	\end{tabular}
\end{table}

\section{Character-level Deep Conflation Models}

\begin{figure}[t]
	\centering
	\includegraphics[width=0.45\textwidth]{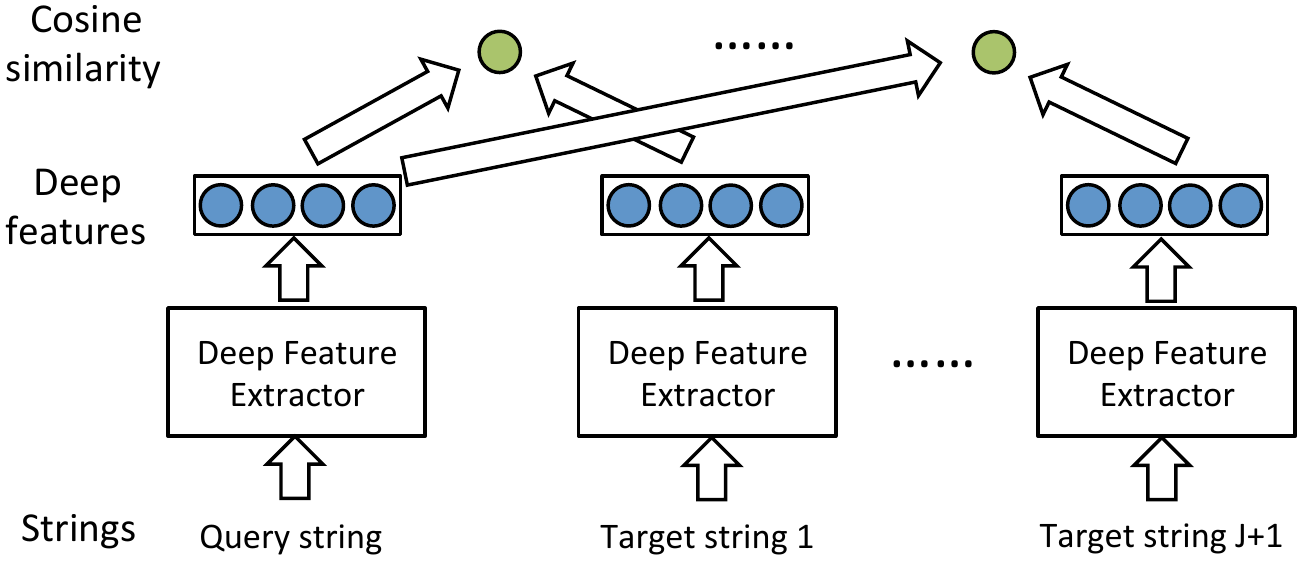}
	\caption{Character-level deep conflation model.}
	\label{Fig:DeepConflation}
\end{figure}

We formulate the deep conflation problem as a ranking problem. That is, given a query string from field A, we rank all the target strings in field B, with the hope that the most similar string in B is ranked on the top of the list. The proposed deep conflation model consists of two parts: (\emph{i}) a deep feature extractor; (\emph{ii}) a ranker. Fig. \ref{Fig:DeepConflation} shows the architecture of the deep conflation model. The deep feature extractors transform the input text strings from character-level into finite dimension feature vectors. Then, the cosine similarity is computed between the query string from field A and all the target strings from field B. The cosine similarity value for each pair of text strings measures the semantic relevance between each pair of the text strings, according to which the target strings are ranked. The entire model will be trained in an end-to-end manner so that the deep feature extractors are encouraged to learn the proper feature vectors that are measurable by cosine similarity. In the rest of this section, we will explain these two components of the deep conflation model with detail.

\subsection{Deep Feature Extractors}

The inputs into the system are text strings, which are sequences of characters. Note that the order of the input characters and words is critical to understand the text correctly. For this reason, we propose to use two deep learning models that are able to retain the order information to extract features from the raw input character sequences. The two deep models we use are: (\emph{i}) Recurrent Neural Networks (RNNs); (\emph{ii}) Convolutional Neural Networks (CNNs). 

RNN is a nonlinear dynamic system that can be used for sequence modeling. However, during the training of a regular RNN, the components of the gradient vector can grow or decay exponentially over long sequences. This problem with \emph{exploding} or \emph{vanishing} gradients makes it difficult for the regular RNN model to learn long-range dependencies in a sequence \cite{pascanu2013difficulty}. A useful architecture of RNN that overcomes this problem is the Long Short-Term Memory (LSTM) structure. 
On the other hand, CNN is a deep feedforward neural network that first uses convolutional and max-pooling layers to capture the local and global contextual information of the input sequence, and then uses a fully-connected layer to produce a fixed-length encoding of the sequence.  In sequel, we first introduce LSTM, and then CNN. 

\subsubsection{LSTM feature extractor}
\begin{figure}[t]
	\centering	
	\includegraphics[width=0.45\textwidth]{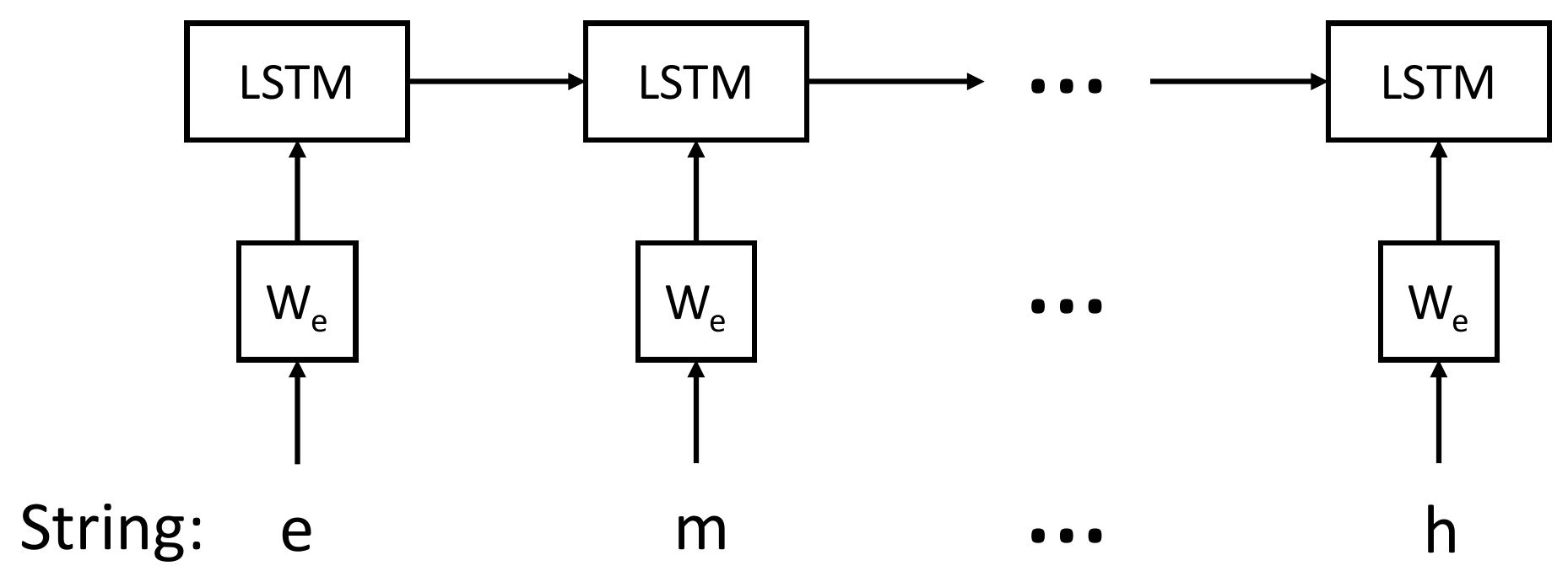} 
	\caption{LSTM based deep feature extractor.}
	\label{fig:lstm}	
\end{figure}

The LSTM architecture \cite{hochreiter1997long} addresses the problem of learning long-term dependencies by introducing a \emph{memory cell}, that is able to preserve the state over long periods of time. Specifically, 
each LSTM unit has a cell containing a state $\cv_t$ at time $t$. This cell can be viewed as a memory unit. Reading or writing the memory unit is controlled through sigmoid gates: input gate $\iv_t$, forget gate $\fv_t$, and output gate $\ov_t$. The hidden units $\hv_t$ are updated as follows:
\begin{align}
\iv_t &= \sigma (\Wmat_{i}\xv_t + \Umat_{i}\hv_{t-1} + \bv_i)\,, 
\label{Equ:Model:LSTM:it}
\\
\fv_t &= \sigma (\Wmat_{f}\xv_t + \Umat_{f}\hv_{t-1} + \bv_f)\,, \\
\ov_t &= \sigma (\Wmat_{o}\xv_t + \Umat_{o}\hv_{t-1} + \bv_o)\,, \\
\tilde{\cv}_t &= \tanh (\Wmat_{c}\xv_t + \Umat_{c}\hv_{t-1} + \bv_c)\,, \\
\cv_t &= \fv_t \odot \cv_{t-1} + \iv_t \odot \tilde{\cv}_t\,, \\
\hv_t &= \ov_t \odot \tanh(\cv_t)\,,
\label{Equ:Model:LSTM:ht}
\end{align} 
where $\sigma(\cdot)$ denotes the logistic sigmoid function, and $\odot$ represents the element-wise multiply operator. $\Wmat_i$ $\Wmat_f$, $\Wmat_o$, $\Wmat_c$, $\Umat_i$, $\Umat_f$, $\Umat_o$, $\Umat_c$, $\bv_i$, $\bv_f$, $\bv_o$ and $\bv_c$ are the free model parameters to be learned from training data.

Given the text string $\qv=[\qv_1,\ldots,\qv_T]$, where $\qv_t$ is the one-hot vector representation of character at position $t$ and $T$ is the number of characters, we first embed the characters into a vector space via a linear transform $\xv_t=\Wmat_e\qv_t$, where $\Wmat_e$ is the embedding matrix. Then for every time step, we feed the embedding vector of characters in the text string to LSTM:
\begin{align}
\xv_t &= \Wmat_e \qv_t, t \in \{1,\ldots,T\}\,, \\
\hv_t &= \mbox{LSTM} (\xv_t), t\in\{1,\ldots,T\}\,,
\end{align}
where the operator $\mathrm{LSTM}(\cdot)$ denotes the operations defined in \eqref{Equ:Model:LSTM:it}-\eqref{Equ:Model:LSTM:ht}. For example, in Fig.~\ref{fig:lstm}, the string $\mathtt{emilio}$ $\mathtt{yentsch}$ is fed into the LSTM. The final hidden vector is taken as the feature vector for the string, i.e., $\yv=\hv_T$. We repeat this process for the query text and all the target texts so that we will have $\yv_Q$ and $\yv_{D_j}$ ($j=1,\ldots J+1$), which will be fed into the ranker to compute cosine similarity (see Sec. \ref{Sec:Ranker}). 

In the experiments, we use a bidirectional LSTM to extract sequence features, which consists of two LSTMs that are run in parallel: one on the input sequence and the other on the reverse of the input sequence. At each time step, the hidden state of the bidirectional LSTM is
the concatenation of the forward and backward hidden states.

\subsubsection{CNN feature extractor}
\label{Sec:Model:CNN}
\begin{figure}[t]
	\centering	
	\includegraphics[width=0.45\textwidth]{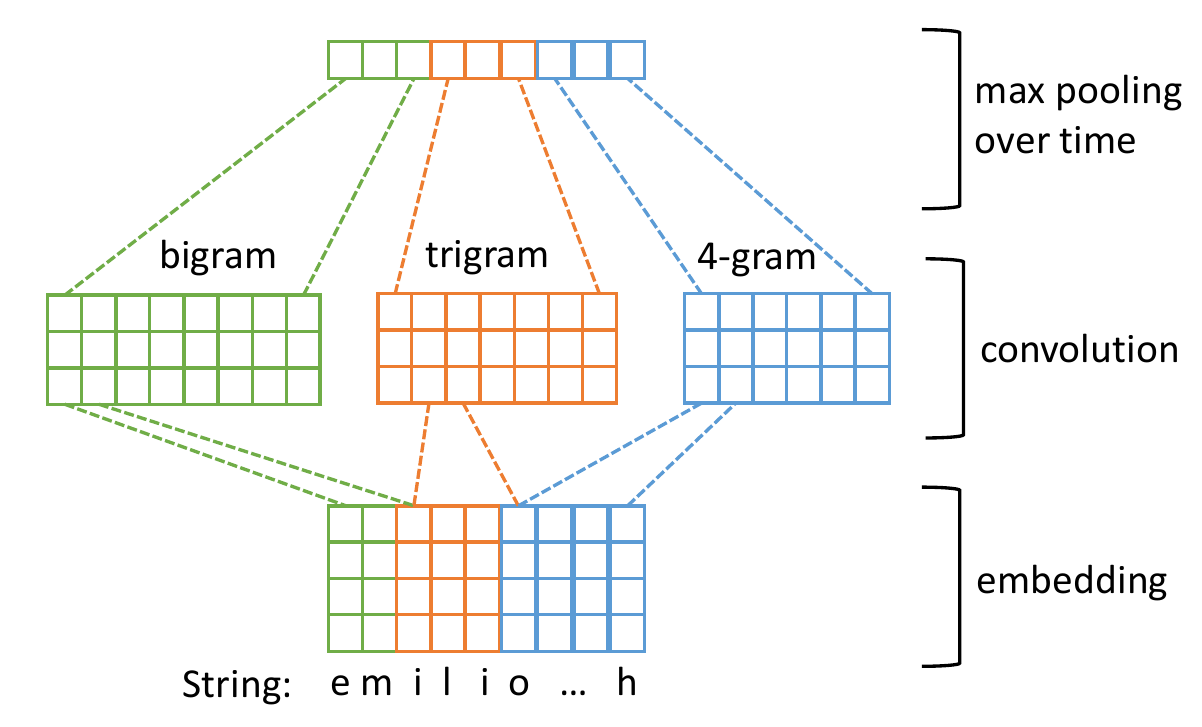} 
	\caption{CNN based deep feature extractor.}
	\label{fig:cnn}	
\end{figure}
Next, we consider the CNN for string feature extraction. Similar to the LSTM-based model, we first embed characters to vectors $\xv_t=\Wmat_e\qv_t$ and then concatenating these vectors:
\begin{align}
\xv_{1:T} = [\xv_1,\ldots,\xv_T]\,.
\end{align}
Then we apply convolution operation on the character embedding vectors. We use three different convolution filters, which have the size of two (bigram), three (trigram) and four (4-gram), respectively. These different convolution filters capture the context information of different lengths. The $t$-th convolution output using window size $c$ is given by 
\begin{align}
\hv_{c,t} = \tanh (\Wmat_c \xv_{t:t+c-1} + \bv_c)\,,
\end{align}
where the notation $\xv_{t:t+c-1}$ denotes the vector that is constructed by concatenating $\xv_t$ to $\xv_{t+c-1}$. That is, the filter is applied only to window $t:t+c-1$ of size $c$. $\Wmat_c$ is the convolution weight and $\bv_c$ is the bias. The feature map of the filter with convolution size $c$ is defined as
\begin{align}
\hv_c = [\hv_{c,1},\hv_{c,2},\ldots,\hv_{c,T-c+1}]\,.
\end{align}
We apply max-pooling over the feature maps of the convolution size $c$ and denote it as
\begin{align}
\hat{\hv}_c = \max \{\hv_{c,1},\hv_{c,2},\ldots,\hv_{c,T-c+1}\}\,,
\end{align}
where the $\max$ is a coordinate-wise max operation. For convolution feature maps of different sizes $c=2,3,4$, we concatenate them to form the feature representation vector of the whole character sequence: $\hv = [\hat{\hv}_2,\hat{\hv}_3,\hat{\hv}_4]\,$. Observe that the convolution operations explicitly capture the local (short-term) context information in the character strings, while the max-pooling operation aggregates the information from different local filters into a global representation of the input sequence. These local and global operations enable the model to encode different levels of dependency in the input sequence. 

The above vector $\hv$ is the final feature vector extracted by CNN and will be fed into the ranker, i.e., $\yv=\hv$.
We repeat this process for the query text and all the target texts so that we will have $\yv_Q$ and $\yv_{D_j}$ ($j=1,\ldots J+1$). The above feature extraction process using CNN is illustrated in Fig.~\ref{fig:cnn}.

There exist other CNN architectures in the literature \cite{kalchbrenner2014convolutional,hu2014convolutional,johnson2014effective}. We adopt the CNN model in \cite{kim2014convolutional,collobert2011natural} due to its simplicity and excellent performance on classification. Empirically, we found that it can extract high-quality text string representations for ranking.

\subsubsection{Comparison between the two deep feature extractors}

Compared with the LSTM feature extractor, a CNN feature extractor may have the following advantages~\cite{gan2016unsupervised}. First, the sparse connectivity of a CNN, which indicates fewer parameters are required, typically improves its statistical efficiency as well as reduces memory requirements. 
Second, a CNN is able to encode regional ($n$-gram) information containing rich linguistic patterns. Furthermore, an LSTM encoder might be disproportionately influenced by characters appearing later in the sequence, while the CNN gives largely uniform importance to the signal coming from each of the characters in the sequence. This makes the LSTM excellent at language modeling, but potentially suboptimal at encoding $n$-gram information placed further back into the sequence.

\subsection{Ranker} 
\label{Sec:Ranker}
Now that we have extracted deep feature vectors $\yv_Q$, $\yv_{D_1}$,..., $\yv_{D_{J+1}}$ from the query and candidate strings, we can proceed to compute their semantic relevance scores by computing their corresponding cosine similarity between query $Q$ and each $j$-th target string $D_j$. More formally, it is defined as
\begin{align}
R(Q,D_j) = \frac{\yv_Q^\top \yv_{D_j}}{||\yv_Q||\cdot ||\yv_{D_j}||}\,,
\end{align}
where $D_j$ denotes the $j$-th target string. At test time, given a query, the candidates are ranked by this relevance scores.

\subsection{Training of the deep conflation model}

We now explain how the deep conflation model could be trained in an end-to-end manner. Given that we have the relevance scores between the query string and each of the target string $D_j$: $R(Q,D_j)$, we define the posterior probability of the correct candidate given the query by the following softmax function
\begin{align}
P(D^{+}|Q) = \frac{\exp(\gamma R(Q,D^{+}))}{\sum_{D^\prime \in \Dmat} \exp(\gamma R(Q,D^\prime))}\,,
\end{align}
where $D^{+}$ denotes the correct target string (the positive sign denotes that it is a positive sample), $\gamma$ is a tuning hyper-parameter in the softmax function (to be tuned empirically on a validation set). $\Dmat$ denotes the set of candidate strings to be ranked, which includes the positive sample $D^{+}$ and $J$ randomly selected incorrect (negative) candidates $\{D_j^{-}; j=1,\ldots,J\}$. The model parameters are learned to maximize the likelihood of the correct candidates given the queries across the training set. That is, we minimize the following loss function
\begin{align}
L(\thetav) = -\log \prod_{(Q,D^{+})} P(D^{+}|Q)\,,
\end{align}
where the product is over all training samples, and $\thetav$ denotes the parameters (to be learned), including all the model parameters in the deep feature extractors. The above cost function is minimized by back propagation and (mini-batch) stochastic gradient descent.

\section{Experimental Results}

\subsection{Dataset}

We evaluate the performance of our proposed deep conflation model on a corporate proprietary business dataset. Since each string can be considered as a sequence of characters, the vocabulary size is $32$ (including one period symbol and one space symbol), which includes the following elements:
$$\mathtt{D M P S a b c d e f g h i j k l m n o p q r s t u v w x y z .}$$
Specifically, the dataset contains $10,000$ pairs of query and the associated correct target string (manually annotated). The average length of the string is $14.47$ with standard deviation $2.89$. The maximum length of the strings is $26$ and the minimum length is $6$. 

\begin{table*}[t]
	\caption{\small{10-fold cross validation results using BoC, LSTM and CNN model, respectively. \textbf{R@K} denotes Recall@K (higher is better). \textbf{Med $r$}, \textbf{Mean $r$} and \textbf{Harmonic Mean $r$} is the median rank, mean rank and harmonic mean rank, respectively (lower is better).} }\label{Table:results}
	\centering
	\small
	\begin{tabular}{c|c|c|c|c|c|c}
		\toprule
		\textbf{Model}& \textbf{R@1} & \textbf{R@3} & \textbf{R@10} & \textbf{Med $r$} & \textbf{Mean $r$} & \textbf{Harmonic Mean $r$}\\
		\midrule
		\multicolumn{7}{l}{\emph{Using correct names to query mis-spelled names}} \\
		\midrule
		BoC &  82.09\scriptsize{$\pm$ 1.59} & 92.30\scriptsize{$\pm$ 0.76} & 96.83\scriptsize{$\pm$ 0.36} & 1.0\scriptsize{$\pm$ 0.0} & 2.380\scriptsize{$\pm$ 0.218} & 1.138\scriptsize{$\pm$ 0.009} \\
		LSTM &  86.66\scriptsize{$\pm$ 0.90} & 95.38\scriptsize{$\pm$ 0.53} & 98.54\scriptsize{$\pm$ 0.20} & 1.0\scriptsize{$\pm$ 0.0} & 1.609\scriptsize{$\pm$ 0.092} & 1.095\scriptsize{$\pm$ 0.007} \\
		CNN & 98.90\scriptsize{$\pm$ 0.18} & 99.97\scriptsize{$\pm$ 0.05} & 100.00\scriptsize{$\pm$ 0.00} & 1.0\scriptsize{$\pm$ 0.0} & 1.012\scriptsize{$\pm$ 0.003} & 1.006\scriptsize{$\pm$ 0.001} \\
		\midrule
		\multicolumn{7}{l}{\emph{Using mis-spelled names to query correct names}} \\
		\midrule
		BoC &  83.56\scriptsize{$\pm$ 1.42} & 93.06\scriptsize{$\pm$ 0.80} & 97.35\scriptsize{$\pm$ 0.27} & 1.0\scriptsize{$\pm$ 0.0} & 2.158\scriptsize{$\pm$ 0.128} & 1.131\scriptsize{$\pm$ 0.011} \\
		LSTM &  87.63\scriptsize{$\pm$ 0.92} & 95.50\scriptsize{$\pm$ 0.45} & 98.67\scriptsize{$\pm$ 0.21} & 1.0\scriptsize{$\pm$ 0.0} & 1.584\scriptsize{$\pm$ 0.055} & 1.088\scriptsize{$\pm$ 0.007} \\
		CNN & 99.25\scriptsize{$\pm$ 0.43} & 99.98\scriptsize{$\pm$ 0.06} & 100.00\scriptsize{$\pm$ 0.00} & 1.0\scriptsize{$\pm$ 0.0} & 1.008\scriptsize{$\pm$ 0.005} & 1.004\scriptsize{$\pm$ 0.002} \\
		\bottomrule
	\end{tabular}
\end{table*}

\subsection{Setup}

We provide the deep conflation results using LSTM and CNN for feature extraction, respectively. Furthermore, we also implement a baseline using Bag-of-Characters (BoC) representation of input text string. This BoC vector is then sent into a two-hidden-layer (fully-connected) feed-forward neural networks. In our experiment, we implement 10-fold cross validation, and in each fold, we randomly select 80\% of the samples as training, 10\% as validation, and the rest 10\% as testing dataset. No specific hyper-parameter tuning is implemented, other than early stopping on the validation set.

For the feed-forward neural network encoder based on the BoC representation, we use two hidden layers, each layer contains 300 hidden units, hence each string is embedded as a 300-dimensional vector.
For LSTM and CNN encoder, we first embed each character into a 128-dimensional vector. Based on this, for the bidirectional LSTM encoder, we further use one hidden layer of 128 units for sequence embedding, hence each text string is represented as a 256-dimensional vector.
For the CNN encoder, we employ filter windows of sizes \{2,3,4\} with 100 feature maps each, hence each text string is represented as a 300-dimensional vector. 

For training, all weights in the CNN and non-recurrent weights in the LSTM are initialized from a uniform distribution in [-0.01,0.01]. Orthogonal initialization is employed on the recurrent matrices in the LSTM \cite{saxe2013exact}. All bias terms are initialized to zero. It is observed that setting a high initial forget gate bias for LSTMs can give slightly better results \cite{le2015simple}. Hence, the initial forget gate bias is set to 3 throughout the experiments. Gradients are clipped if the norm of the parameter vector exceeds 5 \cite{sutskever2014sequence}. The Adam algorithm \cite{kingma2014adam} with learning rate $2\times10^{-4}$ is utilized for optimization. For both the LSTM and CNN models, we use mini-batches of size 100. The hyper-parameter $\gamma$ is set to 10. The number of negative candidates $J$ is set to 50, which are randomly sampled from the rest of the candidate strings excluding the correct one. All experiments are implemented in Theano \cite{bastien2012theano} on a NVIDIA Tesla K40 GPU. For reference, the training of a CNN model takes around 45 minutes to go through the dataset 20 times. 

\begin{table}[t]
	\caption{\small{Average scores for each of the top four retrieved items.} }\label{Table:score}
	\centering
	\small
	\begin{tabular}{c|c|c|c}
		\toprule
		\textbf{top 1} & \textbf{top 2} & \textbf{top 3} & \textbf{top 4} \\ 
		\midrule
		0.792\scriptsize{$\pm$ 0.086} &  0.448\scriptsize{$\pm$ 0.072} & 0.397\scriptsize{$\pm$0.050} & 0.371\scriptsize{$\pm$0.042} \\ 
		\bottomrule
	\end{tabular}
\end{table}

\subsection{Results}
Performance is evaluated using Recall@K, which measures the average times a correct item is found within the top-K retrieved results.
Results are summarized in Table~\ref{Table:results}. 
As can be seen, both of the proposed deep conflation models with LSTM and CNN feature extractors achieve superior performance compared to the BoC baseline. This is not surprising, since sequential order information is utilized in LSTM and CNN.
Furthermore, we observe that CNN significantly outperforms LSTM on this task.  We hypothesize that this observation is due to the fact that the local (regional) sequential order information (captured by CNN) is more important than the gloabl sequential order information (captured by LSTM) in matching two names. For example, if we reverse the family name and given name of a given query name, LSTM might be more prone to mistakenly classifying these two names to be different, while in fact they refer to the same person.

For further analysis, we checked the CNN results on one predefined train/validation/test splits of the dataset. When CNN is used, for Recall@1, out of 1,000 test samples, only 5 samples are mistakenly retrieved. In Table \ref{Table:wrong1}, we show an example of the mistaken case. We can see that the mistakenly retrieved case is quite reasonable. Even humans will make mistakes on these cases. Other four mistakenly retrieved cases are similar and are omitted due to space limit.
The average scores for each of the top four retrieved items are given in Table~\ref{Table:score}. This suggests that, when judging whether two text strings have the same meaning, we can empirically set the threshold to be $(0.792+0.448)/2=0.62$. That is, when the similarity score between two strings is higher than 0.62, we can safely conclude that they refer to the same entity, and we can then conflate the corresponding two rows accordingly.  

\begin{table}[t!]
	\caption{\small{An example of the mistakenly retrieved cases.} }\label{Table:wrong1}
	\centering
	\small
	\begin{tabular}{l|l|l}
		\toprule
		\textbf{query string} & palmer mehaffey & \\
		\textbf{ground truth} & Mr mehaffep paleer & score \\ 
		\midrule
		\textbf{1st result} &   paleer mehaffep & 0.882 \\
		\textbf{2nd result} &   Mr mehaffep paleer & 0.877 \\
		\textbf{3rd result} &    fendlasyn pdlmer &0.427 \\
		\textbf{4th result} &   zalwzar sharley &0.420 \\
		\bottomrule
	\end{tabular}
	\vspace{-0.04cm}
\end{table}

\section{Conclusion}

We have proposed a deep conflation model for matching two text fields in business data analytics, with two different variants of feature extractors, namely, long-short-term memory (LSTM) and convolutional neural networks (CNN). The model encodes the input text from raw character-level into finite dimensional feature vectors, which are used for computing the corresponding relevance scores. The model is learned in an end-to-end manner by back propagation and stochastic gradient descent. Since both LSTM and CNN feature extractors retain the order information in the text, the deep conflation model achieve superior performance compared to the bag-of-character (BoC) baseline.

\bibliographystyle{IEEEbib}
\bibliography{refs}

\end{document}